\pdfoutput=1

\documentclass[a4paper,fleqn, review]{cas-sc}

\usepackage[numbers]{natbib}
\usepackage{subfigure}

\def\tsc#1{\csdef{#1}{\textsc{\lowercase{#1}}\xspace}}
\tsc{WGM}
\tsc{QE}

\begin{document}
\let\WriteBookmarks\relax
\def\floatpagepagefraction{1}
\def\textpagefraction{.001}

\shorttitle{FontTransformer: Few-shot High-resolution Chinese Glyph Image Synthesis via Stacked Transformers}    

\shortauthors{Yitian Liu, Zhouhui Lian}  

\title [mode = title]{FontTransformer: Few-shot High-resolution Chinese Glyph Image Synthesis via Stacked Transformers}  

\author[1]{Yitian Liu}
\ead{lsflyt@pku.edu.cn}

\affiliation[Peking University]{organization={Wangxuan Institute of Computer Technology, Peking University},
            city={Beijing},
            postcode={100871}, 
            country={China}}
            
\author[1]{Zhouhui Lian*}[orcid=0000-0002-2683-7170]
\ead{lianzhouhui@pku.edu.cn}
\ead[url]{https://www.icst.pku.edu.cn/zlian/}

\cortext[1]{Corresponding author}

\begin{abstract}
Automatic generation of high-quality Chinese fonts from a few online training samples is a challenging task, especially when the amount of samples is very small. Existing few-shot font generation methods can only synthesize low-resolution glyph images that often possess incorrect topological structures or/and incomplete strokes. To address the problem, this paper proposes FontTransformer, a novel few-shot learning model, for high-resolution Chinese glyph image synthesis by using stacked Transformers.
The key idea is to apply the parallel Transformer to avoid the accumulation of prediction errors and utilize the serial Transformer to enhance the quality of synthesized strokes. Meanwhile, we also design a novel encoding scheme to feed more glyph information and prior knowledge to our model, which further enables the generation of high-resolution and visually-pleasing glyph images. Both qualitative and quantitative experimental results demonstrate the superiority of our method compared to other existing approaches in few-shot Chinese font synthesis task.
\end{abstract}


\maketitle

\section{Introduction}
Computer fonts are widely used in our daily lives. The legibility and aesthetic of fonts adopted in books, posters, advertisements, etc., are critical for their producers during the designing procedures. Thereby, the demands for high-quality fonts in various styles have increased rapidly. However, font design is a creative and time-consuming task, especially for font libraries consisting of large amounts of characters (e.g., Chinese). For example, the official character set GB18030-2000 consists of 27533 Chinese characters, most of which have complicated structures and contain dozens of strokes~\cite{lian2018easyfont}. Designing or writing out such large amounts of complex glyphs in a consistent style is time-consuming and costly. Thus, more and more researchers and companies are interested in developing systems that can automatically generate high-quality Chinese fonts from a few input samples.

\begin{figure}
	\centering
	\includegraphics[width=1\textwidth]{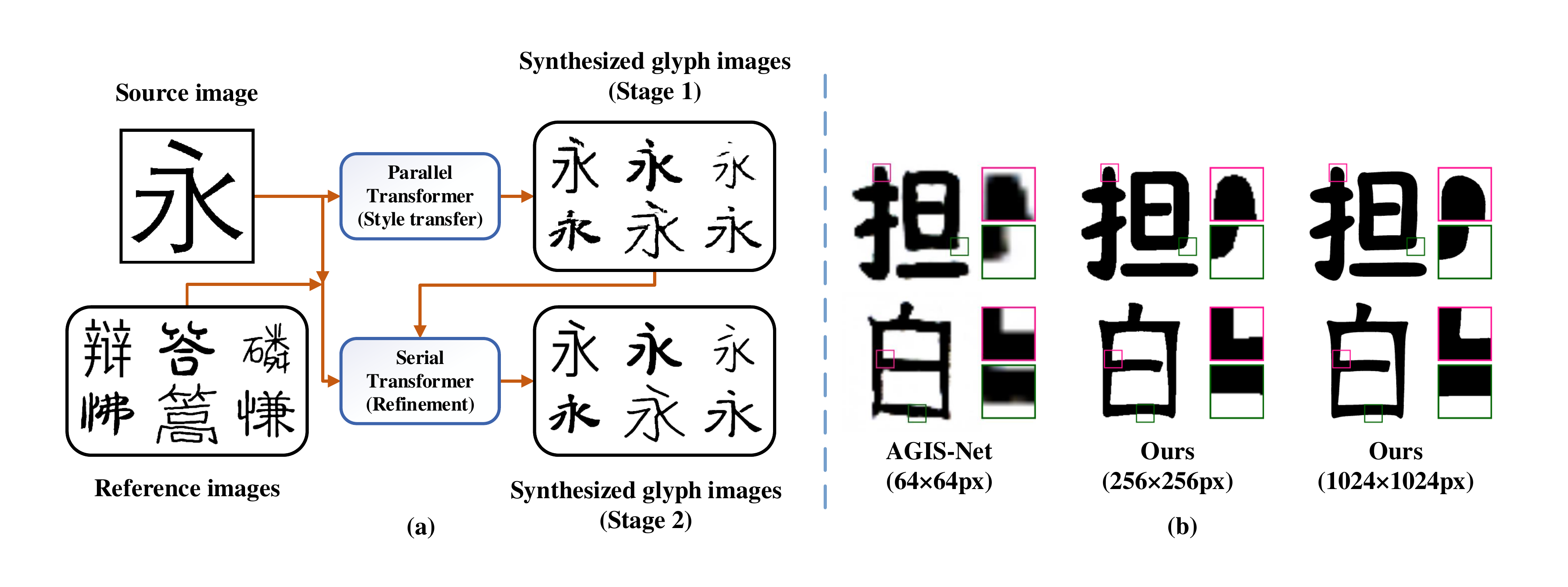}
	\caption{(a) An overview of our method, consisting of a style transfer stage and a refinement stage. (b) Based on few-shot learning, the proposed FontTransformer can synthesize high-resolution (e.g., $1024\times1024$) glyph images needed by font designers to produce high-quality commercial font libraries. Existing approaches (e.g., AGIS-Net) often obtain low-resolution (e.g., $64\times64$) glyph images with blurry outlines.}
	\label{fig:show}
\end{figure}

With the help of various neural network architectures (e.g., CNNs and RNNs), researchers have proposed many DL-based methods for Chinese font synthesis. DL-based methods aim to model the relationship between input and output data (outlines, glyph images, or writing trajectories). Most of them are CNN-based models, such as zi2zi~\cite{tian2017zi2zi}, EMD~\cite{zhang2018separating}, and SCFont~\cite{jiang2019scfont}. Intuitively, we can represent a glyph as the combination of a writing trajectory and a stroke rendering style. Thus, there are some RNN-based methods (e.g., FontRNN~\cite{tang2019fontrnn}) synthesizing the writing trajectory for each Chinese character. Despite the great progress made in the last few years, most existing approaches still need large amounts of offline or online glyph images to train the font synthesis models. Moreover, the quality of vector outlines/glyph images synthesized by those methods is often unsatisfactory, especially when the desired font is in a cursive style or the number of input samples is too small. 

 The design of our FontTransformer is motivated by the observation that a Chinese character is typically rendered as a glyph image when it is composed of sequential strokes in essence. In other words, we can treat a Chinese glyph as an image integrated with serialized information. However, none of the existing methods mentioned above make good use of this feature, leading to fatal problems in the few-shot Chinese font synthesis task: generating glyph images with broken strokes and noisy boundaries. To address this issue, we propose a novel end-to-end few-shot learning model, FontTransformer, which can synthesize high-quality Chinese fonts (see Figure~\ref{fig:show}) from just a few online training samples. Specifically, the proposed FontTransformer is a two-stage model that consists of stacked Transformers, including a parallel Transformer and a serial Transformer. The key idea is to apply the parallel Transformer to avoid the accumulation of prediction errors and utilize the serial Transformer to enhance the quality of strokes synthesized. In this way, our model can learn both the brush and layout styles of the desired font from just a few input images.

Furthermore, font designers typically need high-resolution glyph images (e.g., $1024\times1024$) to create practical font libraries which consist of vector glyphs that enable arbitrary scaling without quality loss. However, existing font synthesis approaches usually synthesize low-resolution glyph images (up to $256\times256$), mainly due to the exponentially increased regression difficulty and memory requirement. To resolve this problem, we design a chunked glyph image encoding scheme based on the fact that there are many repeated patches in binary glyph images. So our model can synthesize visually pleasing and high-resolution glyph images (see Figure~\ref{fig:show}) without markedly increasing computational cost.

In summary, the major contributions of this paper are threefold:
\begin{itemize}
\item We propose FontTransformer, a novel few-shot Chinese font synthesis model, using stacked transformers to synthesize high-resolution (e.g., $256\times256$ or $1024\times1024$) glyph images. To the best of our knowledge, this is the first work that effectively applies Transformers on the task of few-shot Chinese font synthesis.
\item We design a novel chunked glyph image encoding scheme to encode glyph images into token sequences. With this encoding scheme, our method can synthesize arbitrarily high-resolution glyph images by keeping the length of the token sequence a constant.
\item Extensive experiments have been conducted to demonstrate that our method is capable of synthesizing high-quality glyph images in the target font style from a few input samples, outperforming the state of the art both quantitatively and qualitatively.
\end{itemize}

\section{Related Work}
\subsection{Chinese Font Synthesis}
Font design heavily relies on the personal experience of the designer. Although this process can be done with the help of some font editing software such as FontLab\footnote{https://www.fontlab.com/}, it still takes a lot of time and workload to complete this task. In order to generate font quickly and automatically, Campbell and Kautz~\cite{campbell2014learning} built a generative manifold for several standard fonts and generated new fonts by interpolating existing fonts in a high dimensional space. For Chinese characters, Zong and Zhu~\cite{zong2014strokebank} proposed StrokeBank and built a component mapping dictionary from a seed set using a semi-supervised algorithm. The main limitation of StrokeBank is that it is hard to extract perfect strokes or radicals from complex glyphs, especially for those in handwritten styles. To generate a high-quality handwritten font library, Lian et al.~\cite{lian2018easyfont} proposed EasyFont, an automatic system to synthesize personal handwritten fonts by learning style from a set of carefully selected samples.

With the rapid development of deep learning techniques, many DL-based methods have been proposed for the Chinese font synthesis task, such as zi2zi~\cite{tian2017zi2zi}, DCFont~\cite{jiang2017dcfont}, CalliGAN~\cite{wu2020calligan} and so on. Rewrite~\cite{tian2016rewrite} attempted to convert the style of a given glyph image from the source font to the target by using multiple convolution layers. After that, Tian~\cite{tian2017zi2zi} designed zi2zi, a variant of Pix2pix~\cite{isola2017image}, which is capable of synthesizing multiple fonts using a single model by adding the font style embedding. To synthesize more realistic glyph images, Jiang et al.~\cite{jiang2017dcfont} proposed an end-to-end system, DCFont, which only needs 775 or fewer glyph images as training data to learn the style feature. Compared to zi2zi, DCFont can synthesize more visually-pleasing glyph images in the desired font style as input samples, especially for handwritten fonts.

Since the geometric structures of many Chinese characters are complex, existing font synthesis methods often fail to ensure the structure correctness of synthesized glyphs. To solve this problem, some researchers sought the help of prior knowledge of Chinese characters. For instance, SCFont~\cite{jiang2019scfont} and FontRL~\cite{liu2021fontrl} utilized glyph skeletons. Those glyph skeletons contain rich but \emph{expensive} prior knowledge, which helps these two models generate correct glyphs and synthesize high-quality Chinese glyph images. However, it is time-consuming to manually annotate the glyph skeleton for every character in the training dataset. One possible way to handle this problem is to extract glyph skeletons automatically, such as ChiroGAN~\cite{gao2020gan}. In ChiroGAN, Gao and Wu~\cite{gao2020gan} proposed a fast skeleton extraction method (ENet) to replace the manual annotation process. There also exist some methods that employ more \emph{convenient} prior knowledge as guidance. SA-VAE~\cite{sun2017learning} combined the content feature extracted by the Content Recognition Network with the character's encoding, a 133-bits vector. And the encoding vector consists of the information of structure, radical, and character indices. Their experimental results demonstrated that the designed encoding scheme works better than simple character embedding. Similar to SA-VAE, Stroke-GAN~\cite{zeng2021strokegan} used the one-bit stroke encoding to refine the label at the stroke level and used a stroke-encoding reconstruction loss to synthesize better details. Meanwhile, ChinFont~\cite{gao2019automatic}, a system to synthesize vector fonts, introduced the well-known \emph{wubi} coding to represent content information for Chinese characters.

These methods aim to synthesize glyph images in the desired font style. And we can represent the writing trajectory of a Chinese character as a sequence of points. Thereby, some researchers tried to apply the Recurrent Neural Network (RNN) or Long Short-Term Memory (LSTM) models to handle the Chinese font synthesis task. Ha~\cite{ha2015chinese} tried to synthesize writing trajectories by using RNNs for the first time. Zhang et al.~\cite{zhang2017drawing} adopted RNNs to recognize and draw Chinese characters and proposed a pre-processing algorithm to process natural handwritten sequences into model-friendly data. More recently, Tang et al.~\cite{tang2019fontrnn} proposed FontRNN, an RNN-based model with the monotonic attention mechanism and the transfer learning strategy. FontRNN generates the writing trajectories of characters instead of the bitmap images and uses a simple CNN network to synthesize shape details.

\subsection{Few-shot Chinese Font Synthesis}
Few-shot font synthesis aims to synthesize glyph images in the designed font style with very few online training samples, which can further reduce the designer's workload and make it more applicable in some special scenarios, such as ancient books restoration. There exists many few-shot font synthesis methods for Latin or Arabic letters~\cite{azadi2018multi, elarian2015arabic}. Since Chinese consists of a large number of characters and complex glyphs, the performance of these methods on the few-shot Chinese font synthesis task is unsatisfactory. To address this challenging problem, Zhang et al.~\cite{zhang2018separating} proposed EMD, which can handle novel styles with a few reference images. Afterwards, Gao et al.~\cite{gao2019artistic} designed a few-shot artistic glyph image synthesis method with shape, texture, and local discriminators. Zhu et al.~\cite{zhu2020few} proposed a novel method, assigning features as the weights by calculating the deep feature similarity between the target character and reference characters. Then they decoded the weighted feature to the target image. To synthesize more refined details, Huang et al.~\cite{huang2020rd} proposed RD-GAN to crop the output images to many radical images by using the proposed radical extraction module (REM). Then they fed these images to a multi-level discriminator (MLD) to guarantee the global structure and local details. Similarly, instead of modeling global styles, Park et al.~\cite{park2020few} introduced a novel approach by learning localized styles to synthesize glyph images. They decomposed Chinese characters into 371 components and utilized the factorization modules to reconstruct the character-wise style representations from a few reference images. More recently, Liu et al.~\cite{liu2022xmp} proposed XMP-Font adding stroke-level feature to synthesize better inter-component spacing and connected-strokes. Meanwhile, some researchers tried to combine the online data and the offline data to address the few-shot problem by using some unsupervised font generation methods, such as ZiGAN~\cite{wen2021zigan} and DG-Font~\cite{xie2021dg}. Wen et al. proposed ZiGAN, an end-to-end Chinese calligraphy font generation framework utilizing many unpaired glyph images to align the feature distributions. As for DG-Font, its key idea is the proposed Feature Deformation Skip Connection (FDSC) module, which adopts the deformable convolution to perform a geometric transformation on the low-level feature.

Since most writing systems only consist of a small number of characters (e.g., English and Bangla), a similar task is to transfer font styles from other languages to Chinese with a few reference glyph images. FTransGAN~\cite{li2021few} used multi-level attention to extract global and local style features with a few English samples. Park et al.~\cite{park2021multiple} proposed MX-Font to extract style features with a multiple localized experts method in local regions, showing strong generalizability to unseen languages.

\subsection{Vision Transformer}
Due to the utilization of the multi-head self-attention module, Transformers~\cite{vaswani2017attention} have obtained state-of-the-art performance on many sequence processing tasks, such as machine translation~\cite{ott2018scaling}, text generation~\cite{koncel2019text}, document classification~\cite{pappagari2019hierarchical}, question answering~\cite{shao2019transformer, gao2021generalized}, text recognition~\cite{kang2022pay, rouhou2022transformer}, and so on. To apply Transformers in computer vision tasks, we can replace some components of CNNs with attention modules. By adding self-attention modules, Zhang et al.~\cite{zhang2019self} proposed SA-GAN, which can generate image details by using features in all locations. Besides, researchers have also tried to add CNN modules/structures into Transformers, such as BoTNet~\cite{srinivas2021bottleneck}, LocalViT~\cite{li2021localvit}, CvT~\cite{wu2021cvt}, and so on. 

On the other hand, we can directly apply Transformers to vision tasks. ViT~\cite{dosovitskiy2020image} showed that the reliance on CNNs in vision tasks is unnecessary, and achieved state-of-the-art performance on the image classification task. Latter, Esser et al.~\cite{esser2020taming} proposed VQGAN, which combines the Transformer and GAN models for high-resolution image synthesis. VQGAN uses GAN to convert an image to a feature sequence and generates new long feature sequences for synthesizing high-resolution images via the Transformer model. To synthesize higher resolution images, VQGAN utilized a Transformer to infer a longer sequence, which inevitably leads to exponentially-increased computational costs. To address this problem, we propose a chunked glyph image encoding scheme and keep the length of the sequence without markedly increasing computational overhead.

Most Transformer-based models need a large-scale database (e.g., ViT~\cite{dosovitskiy2020image} and DETR~\cite{carion2020end}). Thereby, applying Transformers on the few-shot task is now still a challenging and ongoing problem. This paper proposes FontTransformer based on stacked Transformers synthesizing high-quality Chinese glyph images with very few online training samples.

\begin{figure}[t]
	\centering
	\includegraphics[width=\textwidth]{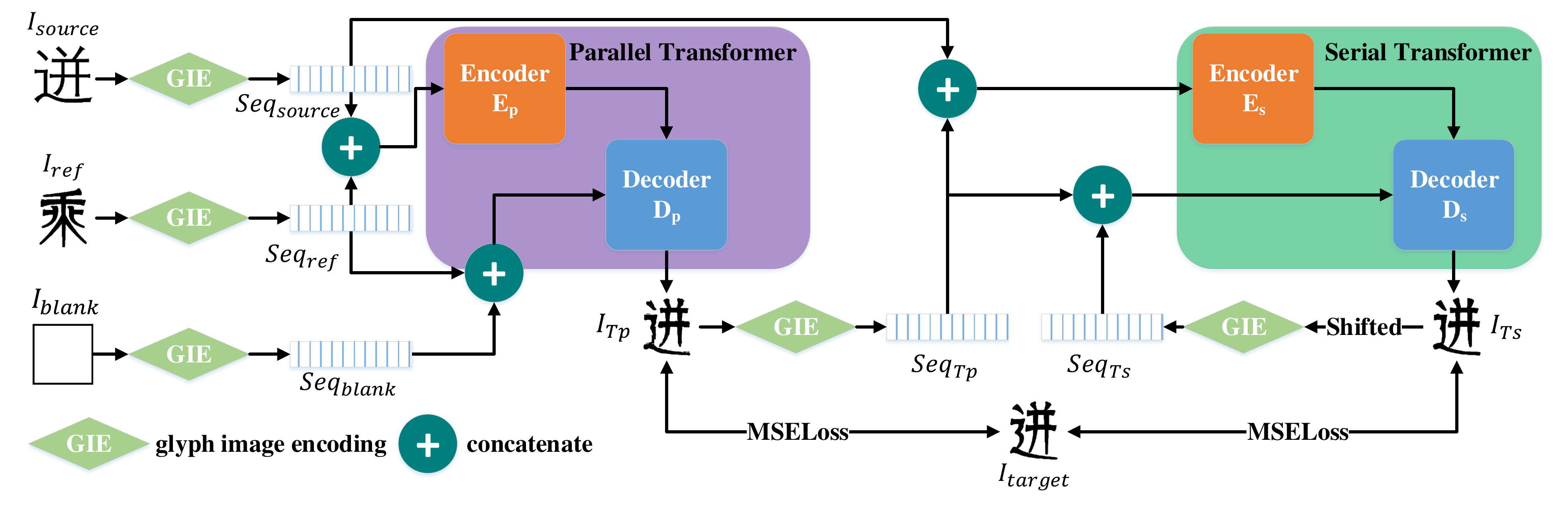}
	\caption{The architecture of our FontTransformer which consists of two parts: the parallel Transformer $T_p$ and the serial Transformer $T_s$. The details of the encoder and decoder are shown in appendix~\ref{appendix:structure}.}
	\label{fig:FontTr}
\end{figure}

\begin{figure}[t]
	\centering
	\includegraphics[width=\textwidth]{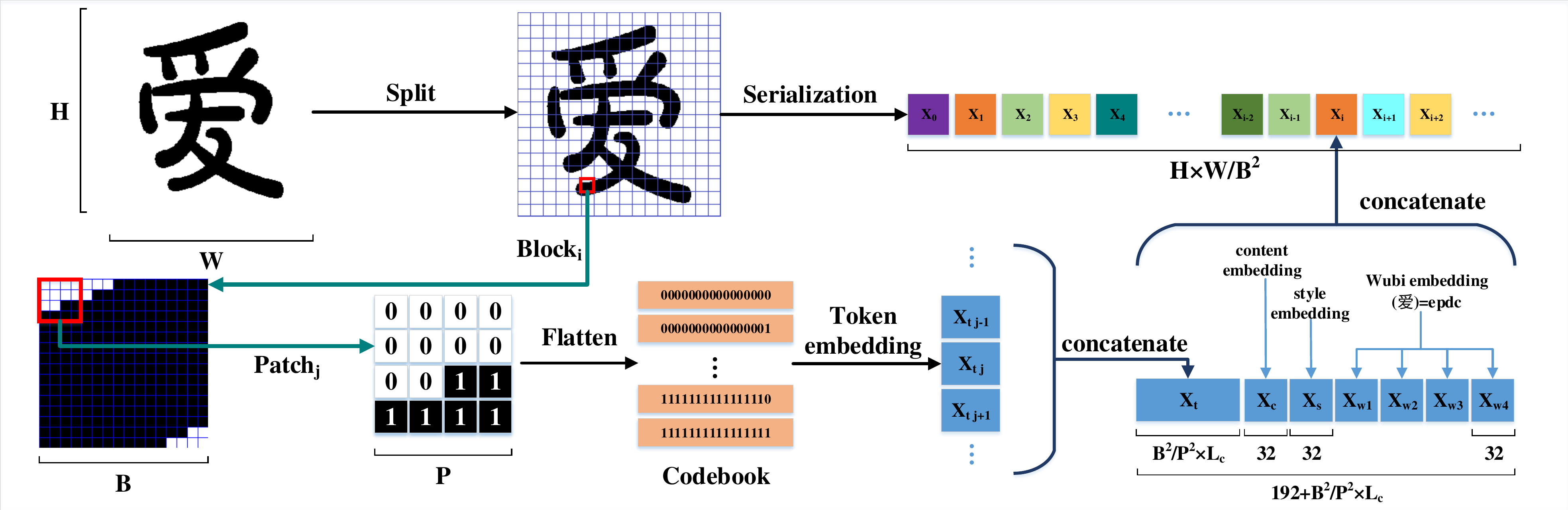}
	\caption{Demonstration of the glyph image encoding procedure in our FontTransformer. We split the image by following the order from left to right, top to bottom. Each patch is serialized to a token embedding $x_t$. Then we combine the token embedding $x_t$, style embedding $x_s$, content embedding $x_c$, and \emph{wubi} embedding $x_w$ into the glyph image encoding vector $x$.}
	\label{fig:encode}
\end{figure}

\section{Method Description}
As shown in Figure~\ref{fig:FontTr}, FontTransformer is a two-stage model with stacked Transformers. In the first stage, we use a Transformer to generate image patches in parallel. Then, to synthesize glyphs with smooth outlines, we sequentially produce image patches by applying another Transformer in the second stage. More specifically, we encode the glyph image to a sequence and add the style, content, and \emph{wubi} embeddings to the sequence in each stage. We will explicitly describe the proposed glyph image encoding and stacked Transformers in the following subsections.

\subsection{Chunked Glyph Image Encoding}
Similar to other vision Transformer methods, we encode glyph images to sequences as the input of our FontTransformer. Since the glyph images are binary bitmaps that can be divided into frequently-repeated patches, we can create a compact sequence to represent a whole glyph image instead of learning complex mappings such as Linear Projection used in ViT~\cite{dosovitskiy2020image} or CNNs in VQGAN~\cite{esser2020taming}.

To reduce the length of sequences, we proposed chunked glyph image encoding (see Figure~\ref{fig:encode}). Specifically, we reshape a single-channel 2D glyph image $I\in\{0,1\}^{H\times W}$ into a sequence of flattened blocks $L=\{I_{b_1}, I_{b_2}, ... ,I_{b_N} | I_{b_i}\in\{0,1\}^{B^2}\}$, where $(H, W)$ denotes the resolution of the glyph image, $(B, B)$ means the size of blocks and $N=\frac{HW}{B^2}$. Since there are many similar parts in the block, we further divide each block $I_b\in\{0,1\}^{B^2}$ into $\frac{B^2}{P^2}$ patches. To be specific, we first flatten the image patch $I_p\in\{0,1\}^{P^2}$ pixel by pixel and express each patch as a binary number $b\in[0, 2^{P^2}-1]$ (see Figure~\ref{fig:encode}). Then we initialize the codebook by sampling $2^{P^2}$ embedding vectors $x_{t_i}\in R^{L_c}$ from the standard normal distribution $N(0,1)$, each corresponding to one binary number. Furthermore, we concatenate the embedding vectors of the corresponding patches into the embedding vector of the block $x_t\in R^{B^2L_c/P^2}$ ($L_c\le P^2$). By adjusting the values of $B$, $P$, and $L_c$, we can apply the above  scheme to encode arbitrarily high-resolution glyph images while constraining the size of the codebook and the length of token, and preserving the information of global structures and local details. We experimentally set the image patch size as $4\times 4$, and thus there are 65536 elements in the learnable codebook. For a $256\times 256$ glyph image, $B$ and $L_c$ are equal to 16 and 16, respectively, and for a $1024\times 1024$ glyph image, they are equal to 64 and 2.

To synthesize high-quality glyph images, we introduce more information and prior knowledge into the glyph image encoding module. Specifically, we add the style embedding $x_s\in[0, 300]$ and the content embedding $x_c\in[1,6763]$ to the token embedding vector\footnote{Our dataset consists of a source font and 300 target fonts, where each font contains 6763 characters. The index number 0 denotes the source font.} $x_t$. To further improve the quality of synthesized images, we combine the token embedding vector with the \emph{wubi}\footnote{Yongmin Wang proposed the \emph{wubi} coding method in 1983, which is designed based on the structure of Chinese characters rather than their pronunciations.} embedding $x_w\in [0,25]^4$, which describes roughly the structure information of a glyph. Similarly, we sample the style, content, and \emph{wubi} embedding vectors from $N(0,1)$. To sum up, as shown in Figure~\ref{fig:encode}, the vector $x$ in the sequence is composed of $x_t, x_s, x_c$, and $x_w$.

\subsection{FontTransformer}
\textbf{Transformer.} The vanilla Transformer~\cite{vaswani2017attention} consists of an encoder and a decoder. The encoder maps the input sequence $X$ to a feature sequence $Z$. Then, the decoder generates an output sequence $Y$ according to $Z$. The key idea of Transformer is the utilization of the attention mechanism, which can be defined as:
\begin{align}
    Attention(Q, K, V) = softmax(\frac{QK^T}{\sqrt{d_k}})V,
\end{align}
where $Q$, $K$, and $V$ are the query, key, and value vectors, respectively, and $d_k$ denotes the dimension of query and key vectors. Our FontTransformer is designed based on the vanilla Transformer with the masked attention mechanism.

\textbf{Stacked Transformers.} Since the decoder of the Transformer sequentially generates tokens, the error accumulation during inference becomes serious and inevitable. This leads to incorrect topological structures and/or incomplete strokes when applying the original Transformer to synthesize glyph images with complicated structures. Specifically, if the start patch of a stroke is missing, the subsequent patches of the entire stroke are hard to be generated correctly. To solve this problem, we propose to use stacked Transformers that consist of a parallel Transformer $T_p$ and a serial Transformer $T_s$.

In the parallel Transformer $T_p$, we use the generative decoder (similar to \emph{Generative Inference} used in Informer~\cite{zhou2020informer}) to simultaneously generate all image patches. The source glyph image sequence $Seq_{source}$, the reference glyph image sequence $Seq_{ref}$, and a blank glyph image sequence $Seq_{blank}$ are fed into the decoder as input. Since there are style, content, and \emph{wubi} embeddings in our glyph image encoding, we feed $Seq_{blank}$ into the decoder instead of only feeding $Seq_{ref}$. It can also keep the architecture of the parallel transformer identical to the serial transformer. However, as shown in Figure~\ref{fig:FontTr}, the glyph image $I_{Tp}$ synthesized in the first stage contains noises and artifacts. Therefore, we adopt the serial Transformer $T_s$ in the second stage to further improve the quality of synthesis results. The serial Transformer $T_s$ generates image patches sequentially based on the synthesis results of the previous stage. The cumulative error still exists in this stage, but the error is tiny and has little influence on the final synthesized glyph images. The two Transformers have similar architectures. They both take the glyph image encoding vector as input and output image patches. The difference between them is whether the patches are generated in parallel or serial during the inference phase.

To sum up, we decompose the Chinese font synthesis task into two steps: style transfer and refinement, and utilize two modified Transformer models to implement them separately. In this manner, our model can synthesize high-quality glyph images with sharp outlines, correct topological structures and free of noises and artifacts.

\textbf{Masked attention.} \label{sec:mask_attention} When calculating the attention function in our model, we add a mask to the function to filter out useless information. The masked attention is defined as:
\begin{align}
    Attention(Q, K, V) = softmax(\frac{QK^T\odot mask}{\sqrt{d_k}})V,
\end{align}
where $x \odot y$ is $-\infty$ if $y$ is 0, otherwise the value is $x$.

After implementing glyph image encoding, each image patch can be represented as a vector. As we can see from Figure~\ref{fig:encode}, blank patches, whose token embedding vectors $x_t$ are $\vec{0}$, cover more than half of the area in the glyph image. Since these image patches contain very few content and style information in the encoder, we set their corresponding mask weights to 0. Intuitively, weights of other patches are set to $1$. In other words, the masks used in the encoders of our two Transformer models can be constructed by:
\begin{equation}
    mask_{encoder} =\left\{
    \begin{aligned}
    1~&~x_t\not=\vec{0}  \\
    0~&~x_t=\vec{0}.
    \end{aligned}
    \right.
\end{equation}
This simple strategy is an effective way to remove tokens with less information and keep the sequence in the same length, thus retaining the position information of each image patch. Similarly, in the parallel Transformer $T_p$, the input of the decoder consists of the source image and the blank image, and we set the mask of the blank image to 0.

\textbf{Loss function.} Since our method generates image patches instead of tokens, we can directly use some loss functions applied in vision tasks, such as the perceptual loss~\cite{johnson2016perceptual} and the contextual loss~\cite{mechrez2018contextual}. However, through our experiments, we find that those losses have little influence on the quality of synthesis results. Finally, we choose to use the MSE Loss which is well suited for our FontTransformer, which can be computed by:
\begin{align}
    Loss_{T_p} = ||I_{T_p}-I_{target}||^2_2; \\
    Loss_{T_s} = ||I_{T_s}-I_{target}||^2_2.
\end{align}
Specifically, our parallel and serial transformers generate a long patch sequence according to the reference and target sequences, so we use the latter half of patches to calculate the loss function.

\section{Experiments}
\subsection{Dataset and Competitors}
In experiments, we compare our FontTransformer with three representative models: zi2zi~\cite{tian2017zi2zi}, EMD~\cite{zhang2018separating}, and DG-Font~\cite{xie2021dg}, and two SOTA few-shot Chinese font synthesis methods: AGIS-Net~\cite{gao2019artistic} and MX-Font~\cite{park2021multiple}. To evaluate the performance of these models, we build a dataset, which consists of 300 fonts. Specifically, each font consists of 6763 glyph images, and we use the Arial Regular font as the source font. Offline, we use 251 fonts (about 1.7 million glyph images) to pre-train models. Online, we use only 100 glyph images to fine-tune those pre-trained models and test them on other 5988 characters in 30 unseen fonts. In Table~\ref{tab:quantitative}, we evaluate and report the results of some SOTA methods on our test set.

\subsection{Implementation Details}
As mentioned above, the proposed FontTransformer can synthesize Chinese glyph images in various resolutions. Our high-resolution (e.g., $1024\times1024$) synthesis results in Figure~\ref{fig:show} and the appendix show the advantage of our method against other existing approaches in image quality. To conduct both qualitative and quantitative experiments for a fair comparison of different methods, we choose to synthesize glyph images in the resolution of $256\times256$ for our FontTransformer. Thus, the block size is $16\times 16$ and each glyph image can be represented as a sequence consisting of 256 tokens through glyph image encoding. The encoder and decoder in two Transformers consist of six MHSA layers and six feed-forward layers, and more details are listed in the appendix. We use the Adam optimizer with $\beta_1=0.9$ and $\beta_2=0.98$ to train our model. The learning rate is $1e-4$ when fine-tuning the online data. During pre-training on online data, we adjust the learning rate dynamically as follows:
\begin{equation}
    \begin{aligned}
        LearningRate(step) = factor\times d_{token} \times min(step^{-0.5}, step\times warmup^{-1.5}),
    \end{aligned}
\end{equation}
where $factor$ is set to 1 and $d_{token}$ denotes the dimension of token. Empirically, we set $d_{token}=448$, $warmup=400$, and the batch size as 64. We train models for 10 epochs in the stage of pre-training and 100 epochs for fine-tuning.

\subsection{Evaluation Metrics}
We adopt three metrics, Mean Absolute Error (MAE), Fréchet Inception Distance (FID)~\cite{heusel2017gans}, and classification accuracy, to evaluate the visual quality of synthesized glyph images. More specifically, MAE (or L1-Loss) mainly reflects the quality of generated images at the pixel level, and FID evaluates the realism and diversity of synthesized images at the feature level. We use the official implementation in PyTorch~\cite{Seitzer2020FID} to calculate FID, which utilizes an Inception-V3 model to extract features. Following DM-Font~\cite{cha2020dmfont} and LF-Font~\cite{park2020few}, we assess the image quality from content-aware and style-aware aspects, respectively. To be specific, we train two VGG-19 networks~\cite{simonyan2014very} on the 300 fonts to classify font styles and character contents, obtaining 99.56\% and 99.87\% classification accuracies, respectively. Then, we classify the glyph images synthesized by different methods using these classifiers and denote the classification accuracies as the Acc(style) and Acc(content) metrics. Also, we calculate the harmonic average of Acc(style) and Acc(content) as the Acc(Hmean) metric to comprehensively evaluate the performance of these methods. We hope that the synthesis glyphs should not only be similar to the target font but also be distinctive from other fonts. So we use the 300 fonts to train two VGG networks instead of only using the 30 fonts in the test set.

\begin{figure}[h]
	\centering
	\subfigure[Brush]{\includegraphics[width=0.3\textwidth]{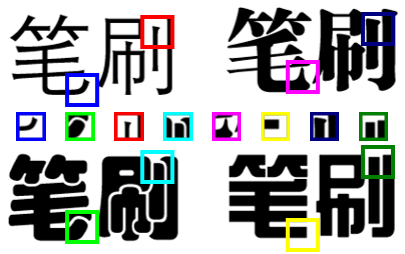}}
	\subfigure[Layout]{\includegraphics[width=0.3\textwidth]{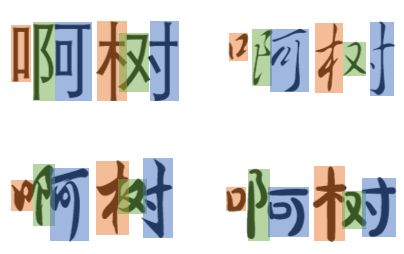}}
	\subfigure[Scale]{\includegraphics[width=0.3\textwidth]{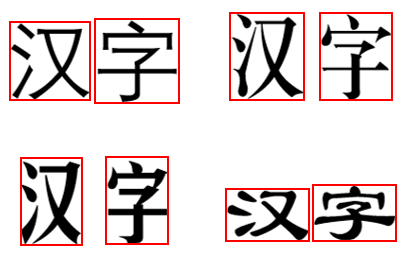}}
	\caption{Three attributes of font style.}
	\label{fig:style}
\end{figure}

\begin{table*}[t]
	\centering
	\caption{Quantitative results of our FontTransformer and other methods.}
	\begin{tabular*}{\tblwidth}{@{}CCCCCC@{}}
		\toprule
		Model & MAE$\downarrow$ & FID$\downarrow$ & Acc(content)\%$\uparrow$ & Acc(style)\%$\uparrow$ & Acc(Hmean)\%$\uparrow$ \\
		\midrule
		\midrule
		zi2zi &  0.1719 & 137.1 & 51.22 & 2.87 & 5.44 \\
		\midrule
        EMD &  0.1538 & 184.5 & 46.63 & 14.68 & 22.33 \\
		\midrule
		AGIS-Net & 0.1300 & 157.7 & \textbf{97.50} & 27.86 & 43.34 \\
		\midrule
 		MX-Font & 0.2115 & 137.4 & 67.24 & 34.16 & 45.30 \\
		\midrule
		DG-Font & 0.1886 & 150.2 & 84.80 & 12.32 & 21.51 \\
		\midrule
		\midrule
		ours w/o chunked encoding & 0.2520 & 186.2 & 0.03 & 0.21 & 0.05\\
		\midrule
	    ours w/o masked attention &  0.1587 & 88.2 & 14.88 & 11.72 & 13.11 \\
		\midrule
		ours (256x256) & \textbf{0.1251} & \textbf{11.0} & 96.45 & \textbf{92.04} & \textbf{94.19} \\
		\midrule
		ours (1024x1024) & 0.1339 & 12.8 & 92.40 & 79.75 & 85.61 \\
		\bottomrule
	\end{tabular*}
	\label{tab:quantitative}
\end{table*}

\begin{figure}[t]
	\centering
	\includegraphics[width=\textwidth]{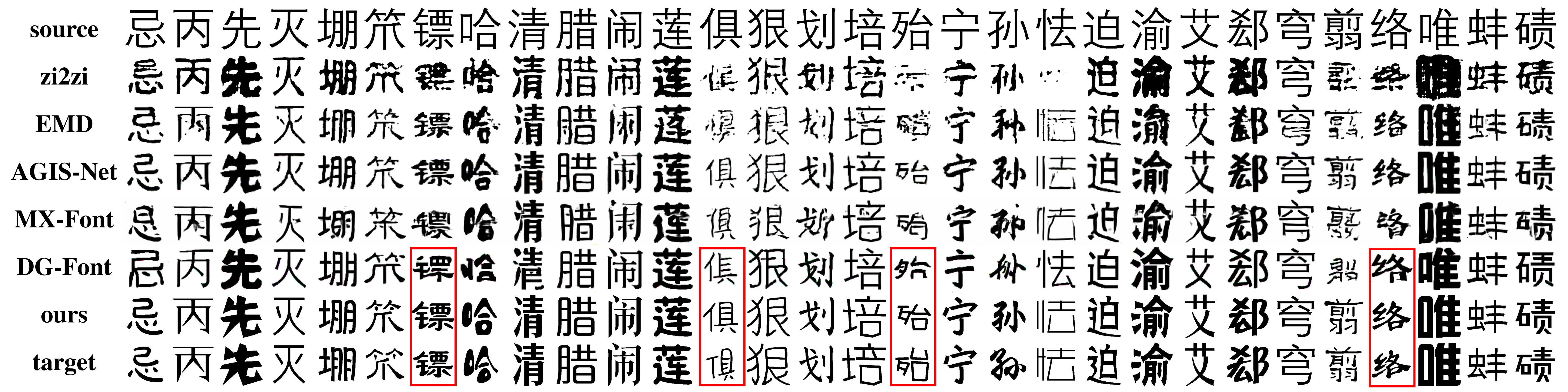}
	\caption{Comparison of synthesis results obtained by our FontTransformer and other five methods. Examples of synthesized glyph images in all 30 test fonts are shown here.}
	\label{fig:visual}
\end{figure}

\subsection{Quantitative and Qualitative Results}
To better describe the quality of images generated by the above-mentioned methods, we roughly decompose the font style into brush, layout, and scale styles. As shown in Figure~\ref{fig:style}, ``Brush'' mainly relates to the rendering effects of stroke trajectories, ``Layout'' describes the locations and structural relationships of strokes or components in a given Chinese character, and ``Scale'' represents the size of a glyph and its location on the canvas. Therefore, two glyph images have the same font style means that their brush, layout, and scale styles are consistent. This also explains why some methods have a low Acc(style) score, although their synthesis results look good.

As shown in Table~\ref{tab:quantitative}, our FontTransformer performs better than other models in most metrics. Figure~\ref{fig:visual} shows some qualitative results of different methods. When only a few (e.g., 100) training samples for fine-tuning are available, zi2zi catches the brush style but synthesizes fuzzy strokes and outlines. Compared with zi2zi, the performance of EMD is better. But EMD outputs glyph images with missing strokes. AGIS-Net can generate glyph images with the target brush, layout, and scale styles. However, there are numerous noises and artifacts in the synthesized images, and some of their strokes are missing. As for MX-Font, it can generate the target brush and the correct layout. But the details in synthesized glyph images with the complex structure are unsatisfactory. Since DG-Font uses unpaired data and adversarial and feature-level loss functions, it often synthesizes glyph images in similar font styles. Specifically, sometimes the brush styles of two synthesized glyph images are similar but not the same as the desired target font. Although DG-Font uses the deformable convolution layers to transfer the glyph structure, there are still some missing strokes when the scales of source and target fonts differ significantly (see red boxes in Figure~\ref{fig:visual}).

We can see that glyph images synthesized by FontTransformer are visually-pleasing and free of noises and missing strokes. Even for the characters with complex structures, our method can synthesize clear and correct glyphs. Moreover, both qualitative and quantitative results demonstrate that our method obtains the best performance in transferring font styles from source to target. More high-resolution glyph images synthesized by our FontTransformer can be found in the appendix. 

\begin{figure}[tp]
	\centering
	\includegraphics[width=1\columnwidth]{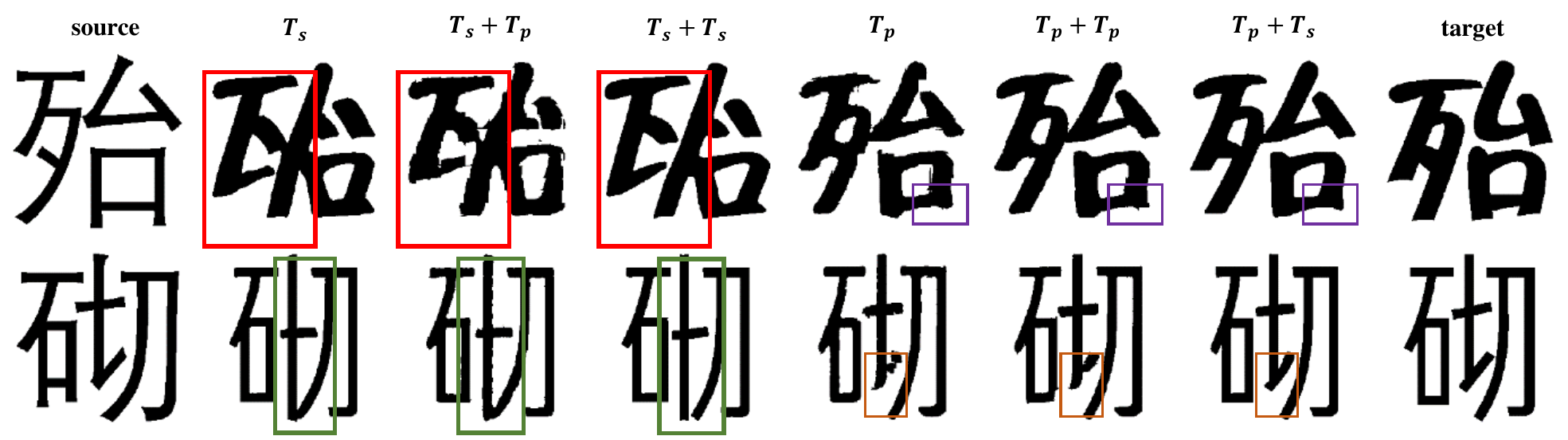}
	\caption{Synthesis results with different combinations of $T_s$ and $T_p$. The parallel Transformer $T_p$ solves the problem of incorrect strokes, and the serial Transformer $T_s$ helps to synthesize glyph images free of noises and artifacts.}
	\label{fig:stacked}
\end{figure}

\subsection{Ablation Study}
\textbf{Stacked Transformers.} As mentioned above, FontTransformer consists of two modules: the parallel Transformer $T_p$ and the serial Transformer $T_s$. From Figure~\ref{fig:stacked}, we can see clearly that the performance of stacked Transformers is better than a single Transformer. It can also be observed that, if we only use the serial Transformer $T_s$, our model often synthesizes incorrect glyphs (the 2nd column in Figure~\ref{fig:stacked}). Moreover, other Transformers can not correct the error (see the third and fourth columns in Figure~\ref{fig:stacked}). On the other hand, if we only use the parallel Transformer $T_p$, there will be noises in synthesized outlines (see the 5th and 6th columns in Figure~\ref{fig:stacked}).

\begin{figure}[tp]
	\centering
	\includegraphics[width=1\columnwidth]{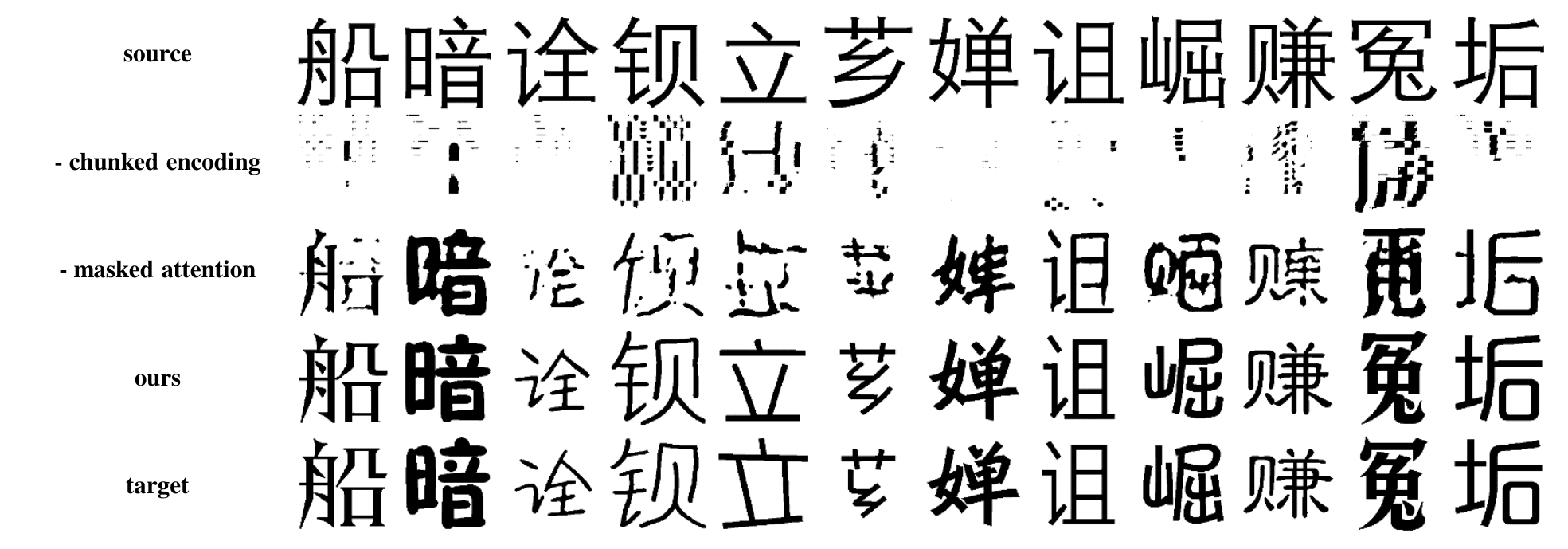}
	\caption{Ablation studies for chunked glyph image encoding and masked attention.}
	\label{fig:mask_encoding}
\end{figure}

\textbf{Chunked glyph image encoding.} Since the glyph images are binary bitmaps, we can employ a compact embedding to represent each patch. Therefore, we use chunked glyph image encoding to avoid more extended token sequences when synthesizing high-resolution glyph images. By adjusting the size of patches and blocks, we can control the length of the token sequences. Furthermore, the learnable codebook helps the model synthesize clear glyph images. As shown in Figure~\ref{fig:mask_encoding}, if we flatten the glyph image as the input without the proposed encoding scheme, it is hard to synthesize the correct glyph.

\textbf{Masked attention.} As mentioned in Section~\ref{sec:mask_attention}, we use masked attention in the encoder of our model. It can filter out most image patches with less information in the image. Figure~\ref{fig:mask_encoding} demonstrates that our FontTransformer without applying the marked attention mechanism results in glyph images with broken/incomplete strokes, while our full model well resolves this problem. 

\begin{figure}[t]
	\centering
	\includegraphics[width=\textwidth]{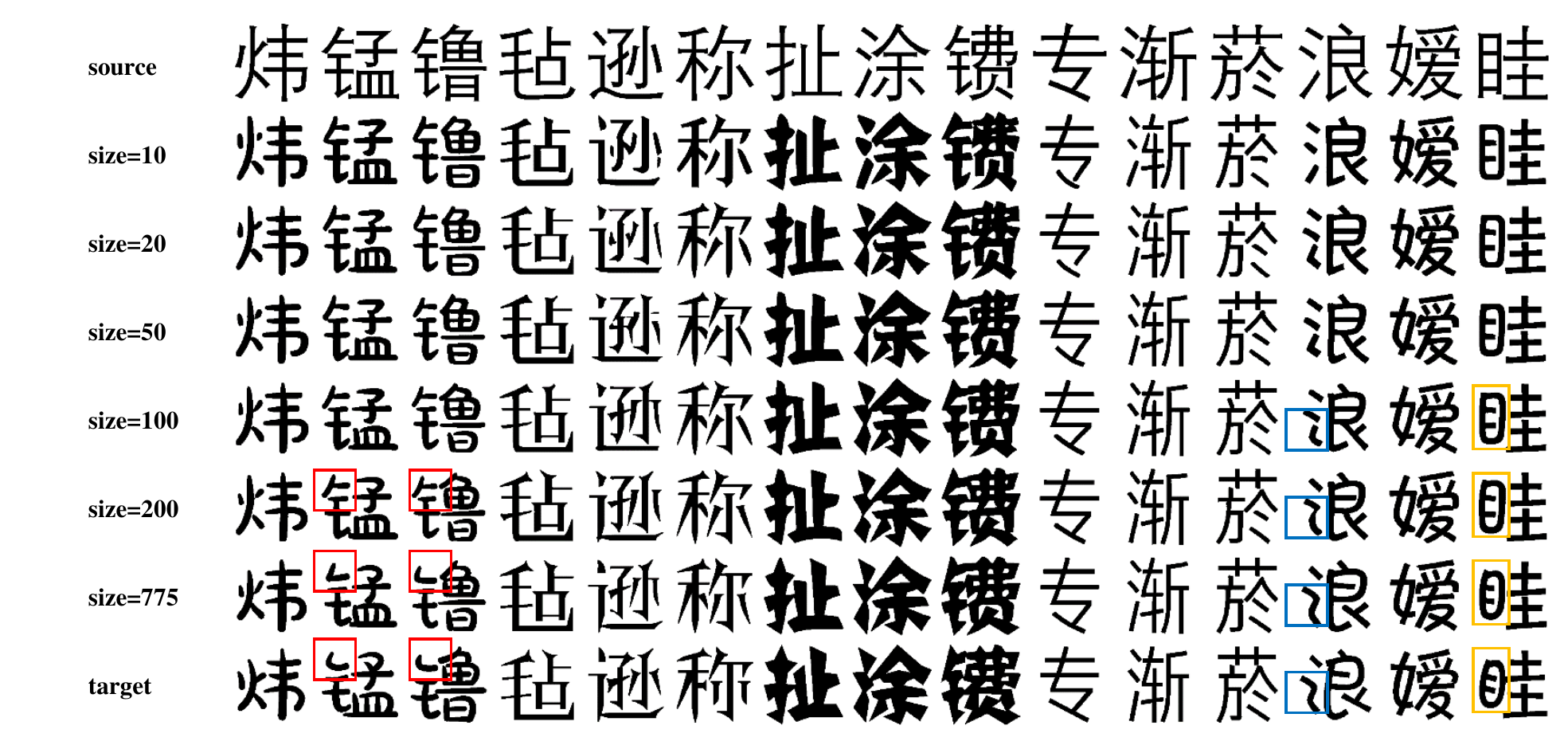}
	\caption{Effects of the number of online data on the performance of FontTransformer. When the few-shot size is 50, synthesis results of our method are already close to satisfactory. We can see that the few-shot size mainly affects the layout style of synthesis results.}
	\label{fig:size}
\end{figure}

\textbf{The number of online data.}
In Figure~\ref{fig:size}, we show the performance of our models with the different numbers of online data. When the number of online data is 10, the model can closely mimic the ``brush'' and ``scale'' styles in the target font, but the layout style is still similar to the source font. Then our synthesis results are already close to satisfactory with 100 online samples. The improvement reflects mainly in the layout style. As the input increases to 200 or 775, the model can ``see'' more radicals. Therefore, the radicals are more similar to the target font (see red boxes in Figure~\ref{fig:size}). We can see that the glyph images synthesized by our method under the conditions of ``size=100'' and ``size=200'' are also satisfactory. What's more, we find that the ``layout'' style is the hardest to transfer. The reasons are two-fold: 1) There is no method to extract the feature of layout from glyph images. 2) The layout is not consistent in some fonts. Some fonts have two or more layout styles for different glyph structures (above to below or left to right). So, how to synthesize glyph images with more similar radical and layout styles as desired but using fewer glyph image samples is still worth exploring in the future.

\begin{figure}[tp]
	\centering	
	\includegraphics[width=1\columnwidth]{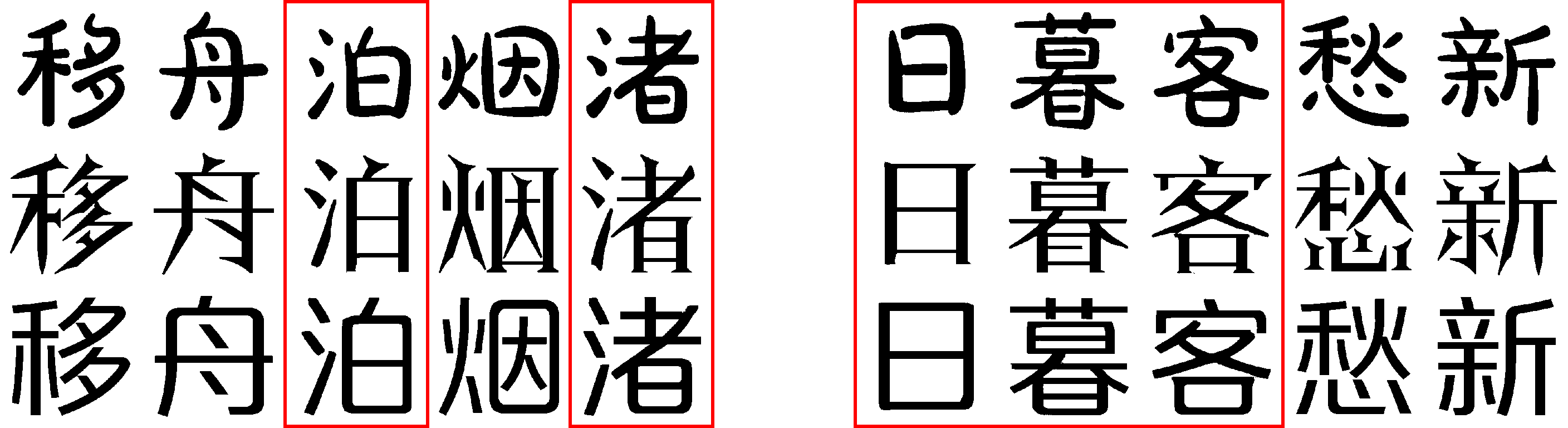}
	\caption{Texts rendered using the synthesized fonts obtained by our method. Machine-synthesized glyphs are marked in red boxes and the others are designed/written by human beings.}
	\label{fig:text}
\end{figure}

\subsection{Text Rendering}
We can consider the quality of synthesis results high if their font style is consistent with the input glyph samples. In Figure~\ref{fig:text}, we render a Chinese poem using the fonts created by our method, where synthesis results are marked in red boxes while the others are written/designed by human beings. We can observe that it is hard to distinguish our machine-generated glyph images from human-created ones, verifying the effectiveness of our method.

\begin{table}[tp]
	\centering
	\caption{User study results.}
	\begin{tabular*}{\tblwidth}{@{}CCCCCCCC@{}}
		\hline
		Font & font-1 & font-2 & font-3 & font-4 & font-5 & font-6 & Average\\
		\hline
		Accuracy & 55.91\% & 53.20\% & 49.36\% & 53.97\% & 54.67\% & 51.33\% & 53.07\% \\
		\hline
	\end{tabular*}
    \label{tab:user_study}
\end{table}

\subsection{User Study}
To further verify the effectiveness of our FontTransformer, we design an online questionnaire. The questionnaire contains six parts, and each part consists of 30 glyph images in the same font style. About half images are synthesized by our method, and the rest are human-created. For each part of the questionnaire, we show five human-created reference images to participants and ask them to pick out all machine-synthesized images from these 30 glyph images shown in the questionnaire. Thirty-one people participated in this user study. In Table~\ref{tab:user_study}, we can see that the average accuracy of participants in each part is close to 50\%. The user study results show the difficulty for human participants to distinguish between our synthesized glyph images from human-created ones, demonstrating the effectiveness of our method. We show the questionnaire in the appendix.

\subsection{Fonts in Dataset}
\begin{figure}[tp]
	\centering	
	\includegraphics[width=0.5\columnwidth]{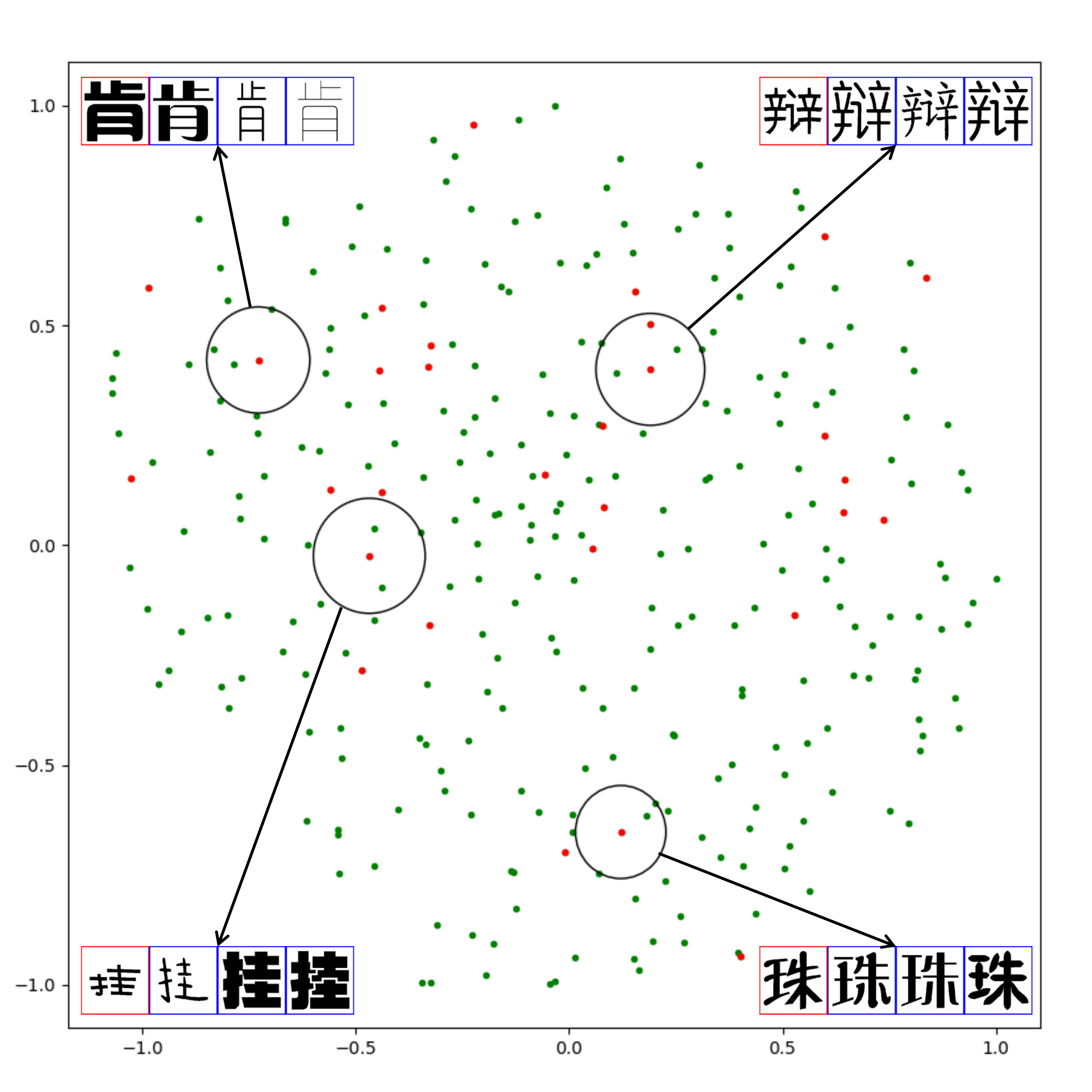}
	\caption{The font style distribution. The red points denote the fonts in the test set, while the others mean the fonts used in the pre-training phase. We show some nearest neighbors of the test fonts.}
	\label{fig:distribution}
\end{figure}

To show the distribution of fonts in our dataset, we use the VGG Network to extract the feature of each font. This VGG network is also used to evaluate the Acc(style). Then we use T-SNE~\cite{maaten2008visualizing} to reduce the feature space dimension and draw them in Fig~\ref{fig:distribution}. To demonstrate that our model does not simply overfit the pre-training set, we also show some nearest neighbor of the fonts used in the evaluation. We can find that the ``brush'', ``layout'', or ``scale'' of neighbor fonts are different from the target font.

\subsection{Limitations}
There still exist some limitations in our proposed FontTransformer, mainly due to the complexity of Transformers. In the inference phase, the parallel Transformer $T_p$ can quickly generate images, but the serial Transformer $T_s$ must generate image patches one by one. Since the length of the sequence and the generation time are linearly related, we use chunked glyph image encoding to constraint the length of the sequence. However, it still takes nearly two hours to synthesize a font with 6763 Chinese characters on a single GTX1080Ti. What is worse, Transformers typically contain more network parameters than CNN-based models. A possible way to solve this problem is to use a more efficient Transformer instead of the original framework. Furthermore, although our method can synthesize high-quality glyph images, there still exist some unsatisfactory synthesis results. Examples of our failure cases can be found in the appendix.

\section{Conclusion and Future Work}
In this paper, we proposed a novel end-to-end few-shot Chinese font synthesis method, FontTransformer. The key idea is to adopt stacked Transformers to synthesize high-resolution glyph images with correct structures and smooth strokes. Specifically, the stacked Transformers consist of a parallel Transformer and a serial Transformer, which are sequentially implemented in two stages. In each stage, we first encoded the images to sequences, which is also integrated with the extra style, content, and \emph{wubi} embeddings. Then, we applied the Transformer to synthesize glyph images in the target font style. Extensive experiments showed that the proposed FontTransformer is capable of synthesizing glyph images with markedly higher visual quality than existing methods in the task of few-shot Chinese font synthesis. In the future, We will try to reduce the complexity of the adopted Transformer models but still further improve the quality of generated images.

\appendix
\section{Questionnaires Used in Our User Study}
\begin{figure}[tp]
	\centering
	\subfigure[A questionnaire in our user study.]{\includegraphics[width=0.45\columnwidth]{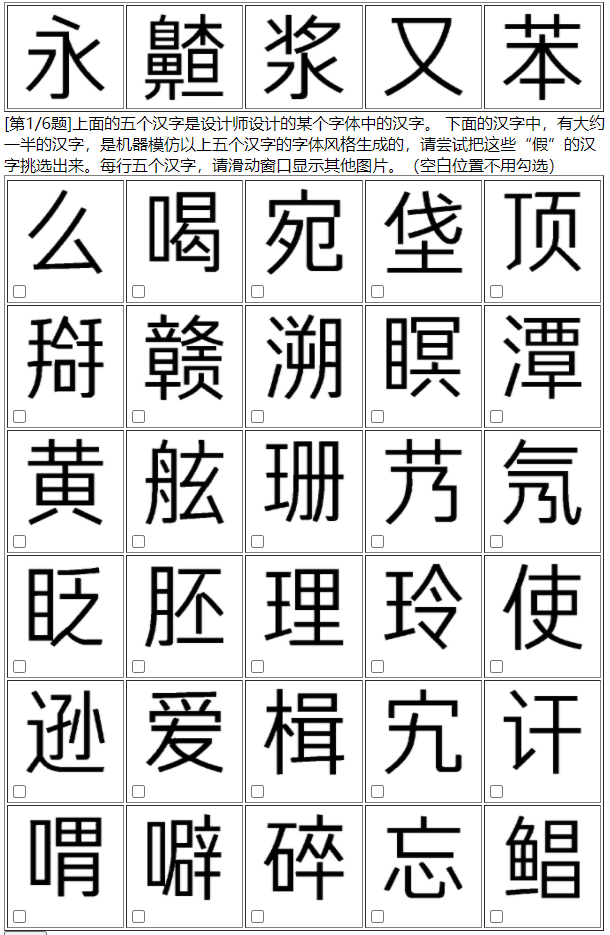}}
	\subfigure[The answer of the left questionnaire. The red labels mark the machine-generated glyph images.]{\includegraphics[width=0.45\columnwidth]{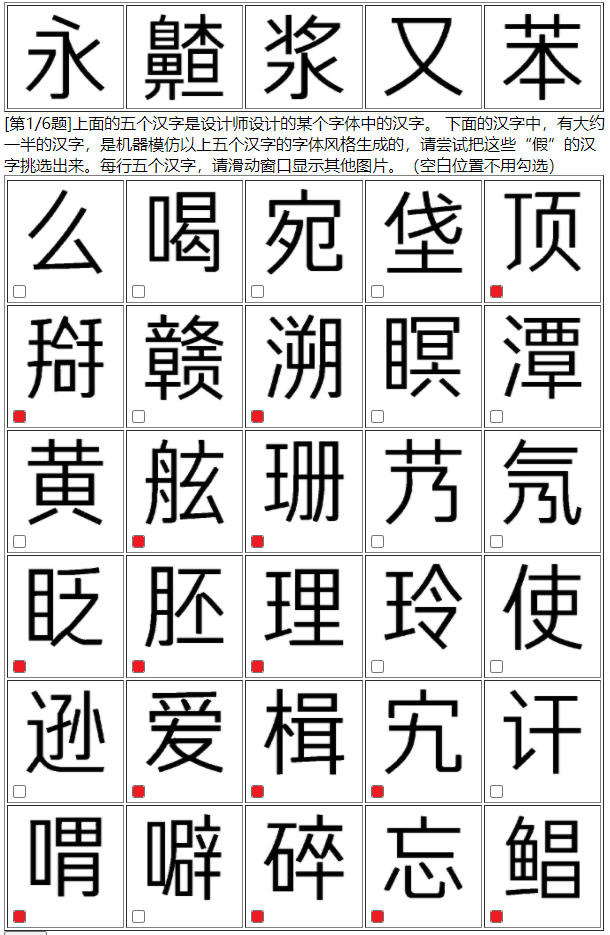}}
	\caption{In the user study, we show each participant six questionnaires. Each questionnaire consists of five human-created images as guidelines (the first lines) and thirty glyph images as questions. The text in the questionnaire means that '[Question 1/6] Above five Chinese characters are designed by a designer. About half of the following Chinese characters are synthesized by the machine (AI) imitating the font style of the above five characters. Please try to select machine-generated characters.'}
	\label{fig:questionnaire}
\end{figure}

In Figure~\ref{fig:questionnaire}, we show a questionnaire of our user study and the corresponding answer. Our experimental results show that it is difficult for participants to pick out machine-generated glyph images from the questionnaires.

\section{Network Architecture}
\label{appendix:structure}
In each stage of FontTransformer, we encode glyph images to sequences and generate image patches in the target font using a Transformer. We can also use other models (e.g., RNN or LSTM) to handle this task. Since Transformers perform well in many vision tasks, we choose to use Transformers as the backbone of our model. In Table~\ref{tab:structure}, we show the network architecture of our FontTransfomer.

\begin{table}[h]
    \centering
    \caption{The network architecture of our FontTransformer.}
    \begin{tabular}{|c|c|}
        \hline
        \multicolumn{2}{|c|}{Encoder}\\ 
        \hline
        Masked Multi-Head Attention & \multirow{4}{*}{$\times 6$} \\
        \cline{0-0}
        Layer-Norm & \\
        \cline{0-0}
        Feed-Forward & \\
        \cline{0-0}
        Layer-Norm & \\
        \hline
        \hline
        \multicolumn{2}{|c|}{Decoder} \\
        \hline
        Multi-Head Self-Attention & \multirow{6}{*}{$\times 6$} \\
        \cline{0-0}
        Layer-Norm & \\
        \cline{0-0}
        Multi-Head Attention & \\
        \cline{0-0}
        Layer-Norm & \\
        \cline{0-0}
        Feed-Forward & \\
        \cline{0-0}
        Layer-Norm & \\
        \hline
        \hline
        \multicolumn{2}{|c|}{Generator} \\
        \hline
        \multicolumn{2}{|c|}{Linear} \\
        \hline
        \multicolumn{2}{|c|}{Tanh} \\
        \hline
    \end{tabular}
    \label{tab:structure}
\end{table}

\begin{figure}[htp]
	\centering
	\subfigure[There are some wrong stroke connections in our synthesized glyph images.]{\includegraphics[width=0.4\columnwidth]{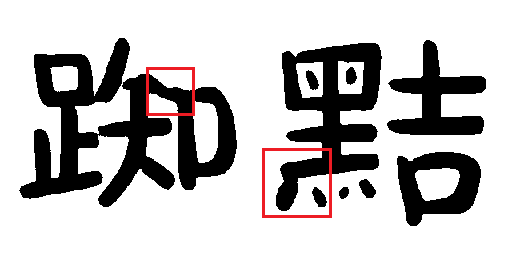}\label{fig:failure_a}}
	\subfigure[The details of the same components in our synthesized glyph images are slightly different.]{\includegraphics[width=0.4\columnwidth]{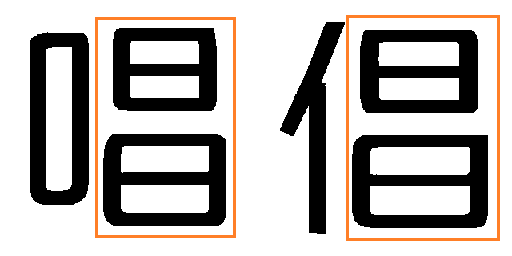}\label{fig:failure_b}}
	\caption{Some failure cases of our FontTransformer.}
	\label{fig:failure}
\end{figure}

\section{Failure Cases}
Experiments demonstrate the performance of our FontTransformer on high-quality glyph image synthesis, but there still exist some failure cases. We show some wrong stroke connections in Figure~\ref{fig:failure_a}. Another problem to be solved is how to keep the consistency of the same components in different characters. As shown in Figure~\ref{fig:failure_b}, the local details of two synthesized glyphs in the orange boxes are inconsistent. In FontTransformer, we only use the pixel-level loss (MSE-Loss) to constrain the synthesis results. If extra adversarial loss functions are applied, we think this problem can be solved to some extent.

\section{More Synthesis results}
\begin{figure}[t]
    \centering
	\includegraphics[width=\columnwidth]{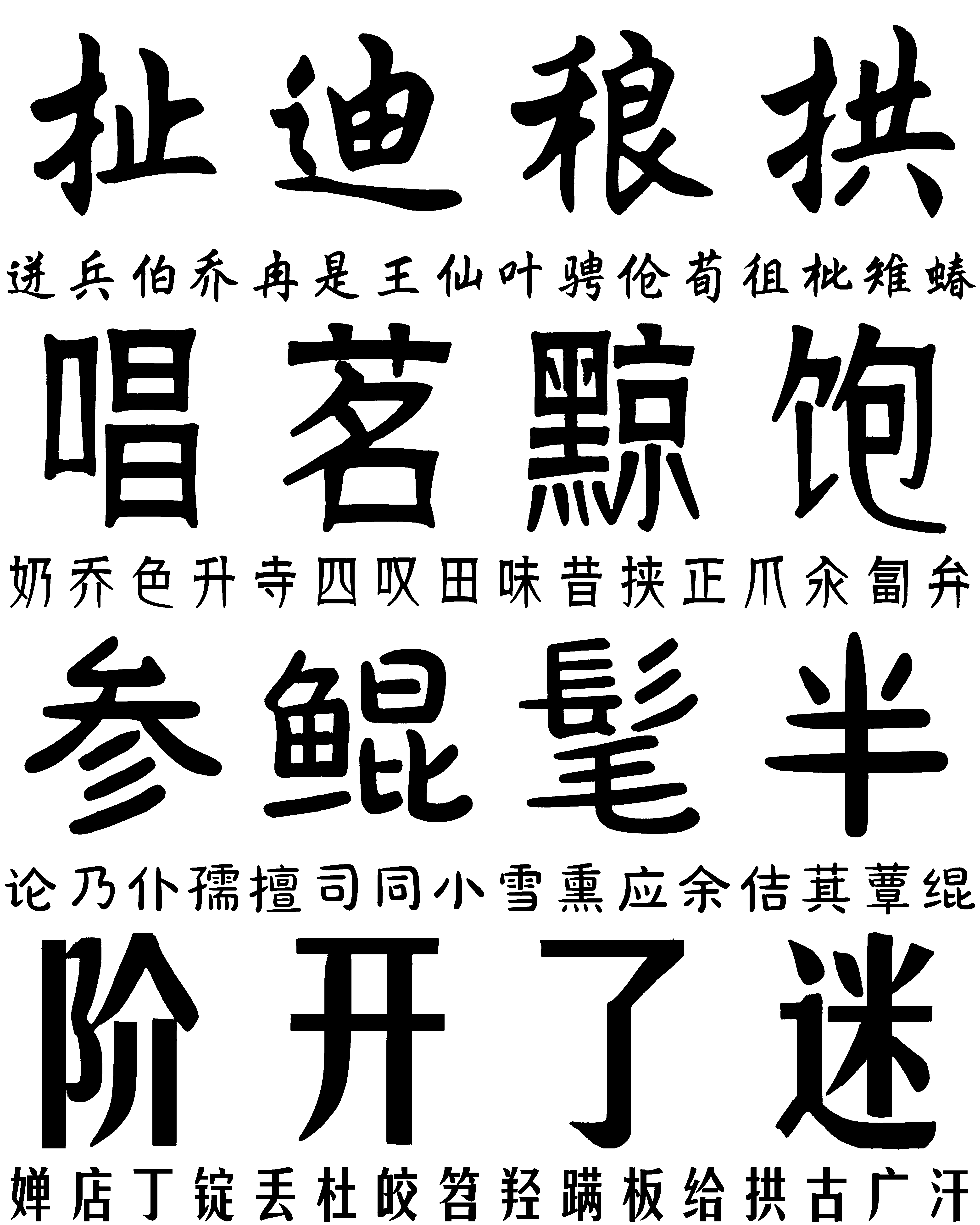}
	\caption{More synthesis results of our FontTransformer, including $1024\times 1024$ glyph images and $256 \times 256$ images.}
	\label{fig:visual_1}
\end{figure}

\begin{figure}[t]
    \centering
	\includegraphics[width=\columnwidth]{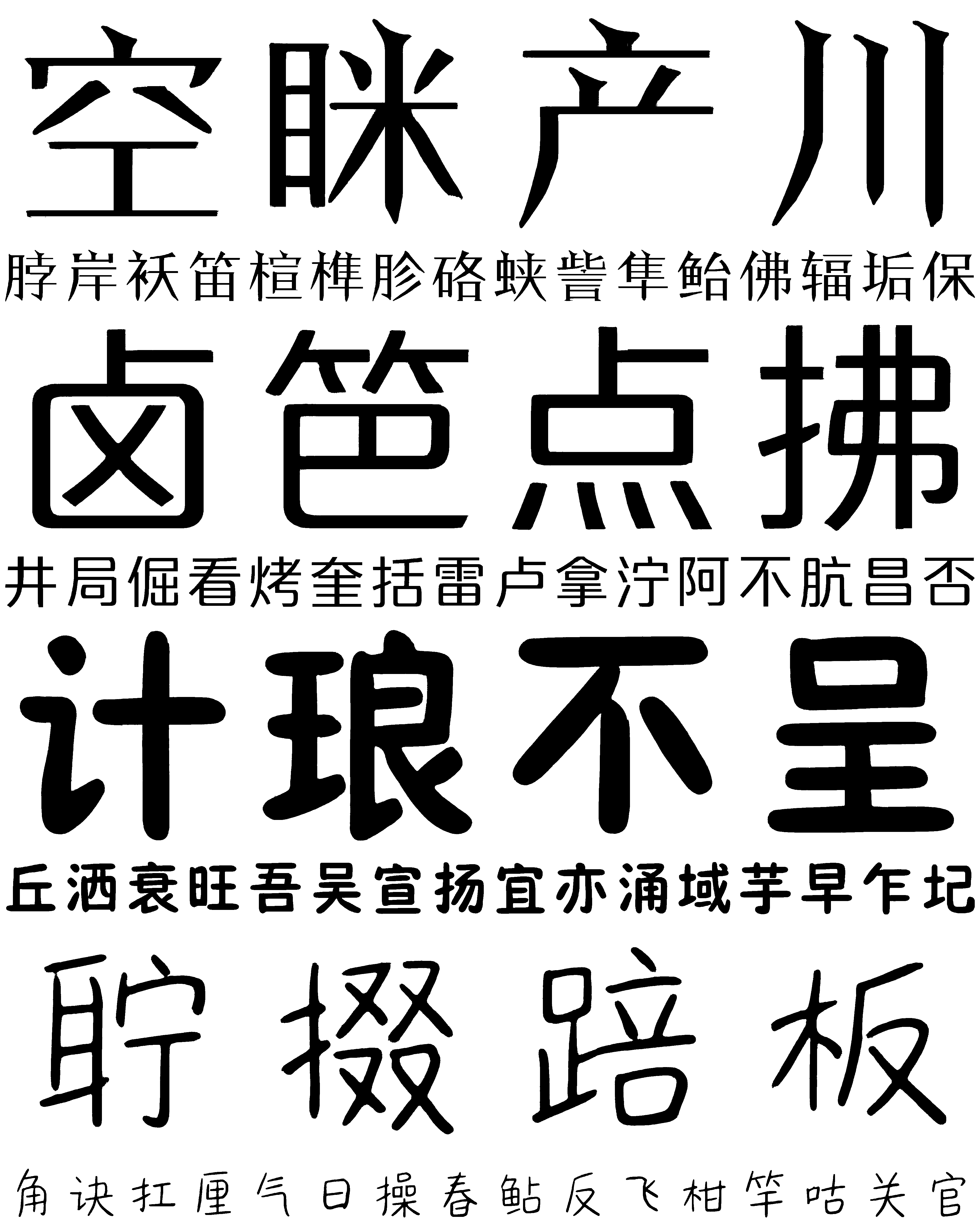}
	\caption{More synthesis results of our FontTransformer, including $1024\times 1024$ glyph images and $256 \times 256$ images.}
	\label{fig:visual_2}
\end{figure}

In Figure~\ref{fig:visual_1},~\ref{fig:visual_2}, we show more synthesis results of our FontTransformer, including the $1024\times 1024$ and $256\times 256$ glyph images.

\printcredits

\bibliographystyle{model1-num-names}

\bibliography{refer}



\end{document}